\documentclass{article}

\PassOptionsToPackage{numbers, compress}{natbib}
\usepackage[main, final]{neurips_2026}

\usepackage[utf8]{inputenc} % allow utf-8 input
\usepackage[T1]{fontenc}    % use 8-bit T1 fonts
\usepackage{hyperref}       % hyperlinks
\usepackage{url}            % simple URL typesetting
\usepackage{booktabs}       % professional-quality tables
\usepackage{amsfonts}       % blackboard math symbols
\usepackage{nicefrac}       % compact symbols for 1/2, etc.
\usepackage{microtype}      % microtypography
\usepackage{xcolor}         % colors
\usepackage{amsmath}
\usepackage{graphicx}
\usepackage{enumitem}
\usepackage{xspace}
\usepackage{array}
\usepackage{longtable}
\usepackage{needspace}

\definecolor{CRGreen}{RGB}{92,171,53}

\newif\ifshowcrchanges
\ifdefined\HBARclean\showcrchangesfalse\else\showcrchangestrue\fi

\showcrchangesfalse

\newcommand{\camera}[1]{{\ifshowcrchanges\color{CRGreen}\fi#1}}
\newcommand*{\modelname}{HBAR\xspace}
\newcommand{\sg}{\operatorname{sg}}
\newcommand{\MSE}{\operatorname{MSE}}
\newcommand{\MAE}{\operatorname{MAE}}

\newcommand{\hbarlow}{HBAR-415}

\newcommand{\scoreci}[3]{%
  \ensuremath{#1}\,{\scriptsize\textcolor{black!55}{(#2, #3)}}%
}

\title{A Hierarchical Tokenization Framework for Voxel-Level fMRI Representation Learning}

\author{%
  Karanpartap Singh\thanks{Correspondence to: \texttt{karanps@stanford.edu}}
  \qquad Yue Zhao
  \qquad Mohammad H. Abbasi
  \qquad Ehsan Adeli \\
  Stanford University
}

\newcommand\blfootnote[1]{%
  \begingroup
  \renewcommand\thefootnote{}%
  \footnotetext{#1}%
  \endgroup
}

\begin{document}

\maketitle

\begin{abstract}
  We introduce \textbf{\modelname} (Hierarchical Brain Activity Representation), a hierarchical tokenization framework for learning compact, multi-scale representations of neuroimaging data. \modelname organizes voxel-level resting-state fMRI into a hierarchy spanning sub-parcel voxel groups, parcels, and large-scale functional networks, enabling \camera{up to} $415\times$ compression while preserving fine-grained spatial information beyond conventional parcellation. The framework supports fixed atlas-based, fully learned, and softly prior-guided hierarchies, allowing us to systematically study the role of structured priors in brain representation learning. On \camera{HCP-YA, at approximately $415\times$ compression, HBAR achieves higher parcel and network correlations than rate-matched flat tokenizers, while at a lower $52\times$ compression rate, shuffling the anatomical grouping reduces parcel correlation from 0.920 to 0.886. Demographic readouts with autoregressive training also reveal potential gains from the hierarchical structure}. By combining \camera{sub-parcel inputs}, compact discrete tokens, and multiscale brain structure, \modelname offers a practical framework for representation learning in high-dimensional neuroimaging data.
\end{abstract}

\begin{figure}[ht]
\centering
\includegraphics[width=\linewidth]{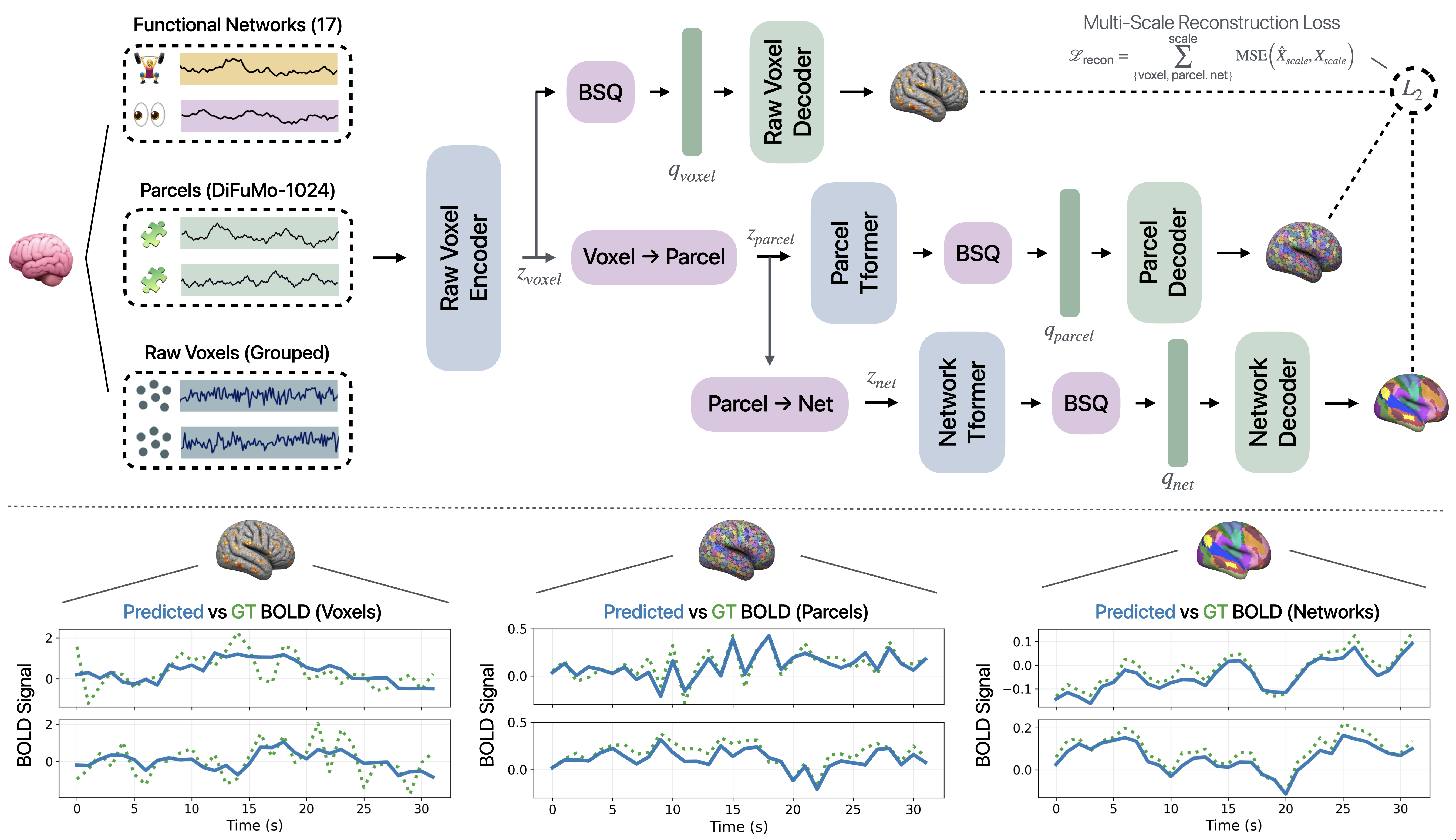}
\caption{\textbf{Hierarchical tokenization and reconstruction for resting-state fMRI.}
Voxel-level BOLD signals are organized into a three-level hierarchy spanning sub-parcel voxel groups, parcels (DiFuMo-1024), and large functional networks (Yeo-17). A shared encoder produces voxel-level representations, which are progressively aggregated to parcel- and network-level tokens via hierarchical mappings. At each scale, tokens are discretized using binary spherical quantization (BSQ) and decoded to reconstruct the corresponding signals. The model is trained using a multi-scale reconstruction objective, encouraging representations to preserve information across voxel, parcel, and network levels. Example reconstructions (bottom) illustrate recovery of BOLD dynamics at all scales.}
\label{fig:intro}
\end{figure}

\section{Introduction}
\label{sec:intro}

Many modern machine learning systems operate on high-dimensional structured data, where observations are defined over sets of interrelated entities rather than independent features. Examples include images defined over spatial grids (e.g. CNNs) \cite{Lecun1998Gradient-basedRecognition}, graphs defined over relational neighborhoods (e.g. Graph-CNNs) \cite{Kipf2017Semi-SupervisedNetworks}, and biological signals defined over anatomically organized domains such as brain networks \cite{ThomasYeo2011TheConnectivity}. A central challenge in these settings is to construct representations that are both compact enough to enable efficient learning and inference, yet expressive enough to preserve the dependencies and organization inherent to the data.\blfootnote{We make our code available at \url{https://github.com/su-karanps/HBAR}.}

A common approach is \emph{tokenization}, which maps high-dimensional inputs into compact representations for downstream modeling. \emph{Discrete} tokenization offers additional benefits: fixed-rate codes provide storage bounds, enabling compression at biobank scale, while finite vocabularies support autoregressive sequence modeling. Existing strategies often rely on fixed partitions or uniform downsampling, which can be suboptimal for structured signals that lack regular spatial organization. An alternative is to view tokenization as learned aggregation, where elements are grouped according to their statistical dependencies \cite{Tian2024VisualPrediction}. This motivates \emph{hierarchical tokenization}, which organizes representations from local to global scales. However, existing approaches often rely on fixed spatial hierarchies (e.g., image pyramids), limiting their applicability to more general structured domains.

Resting-state functional magnetic resonance imaging (rs-fMRI) provides a compelling testbed for hierarchical tokenization. Brain activity is inherently structured across multiple scales, from voxel-level signals to parcels and functional networks, and exhibits rich dependencies that are not strictly localized in space. At the same time, existing representation learning approaches for rs-fMRI typically operate at one of two extremes: either directly modeling voxel-level data at high computational cost \cite{Wang2026TowardsAnalysis}, or aggregating signals into predefined atlases or parcellations \cite{Caro2023BrainLM:Recordings,Dong2024Brain-JEPA:Masking,Yang2024BrainMass:Learning}, which can discard fine-grained spatial information. This tension makes rs-fMRI an ideal setting in which to study learned, multiscale tokenization. To this end, we introduce \textbf{\modelname\ (Hierarchical Brain Activity Representation)}, a multiscale tokenization framework for structured neuroimaging data.

Our contributions are summarized as follows: (1) we introduce a general framework for hierarchical tokenization of structured data based on hierarchical aggregation rather than fixed spatial downsampling; (2) we instantiate \modelname for rs-fMRI using a voxel-to-parcel-to-network hierarchy, achieving \camera{up to} $415\times$ compression \camera{and evaluating a higher-fidelity $52\times$ setting}; (3) we systematically study the tradeoff between fully learned and prior-guided hierarchical organization, including a hybrid approach combining adaptive tokenization with structured neurobiological priors; and (4) we \camera{evaluate} reconstruction and demographic prediction \camera{with repeated training seeds and matched readouts,} finding that atlas-consistent grouping improves parcel reconstruction relative to shuffled grouping and may provide a strong inductive bias for hierarchical representation learning in fMRI.

\section{Related Work}
\label{sec:related_work}

\subsection{Tokenization and Hierarchical Tokenization}

Modern tokenization emerged from variational autoencoders \cite{Kingma2022Auto-EncodingBayes} and VQ-VAE \cite{Oord2018NeuralLearning}, which introduced discrete latent representations compatible with autoregressive modeling. Single-scale discrete tokenizers such as VQGAN \cite{Esser2021TamingSynthesis} and DALL-E \cite{Ramesh2021Zero-ShotGeneration} showed that these codes support large-scale generative modeling, and VQ-VAE-2 \cite{Razavi2019GeneratingVQ-VAE-2} introduced the multiscale formulation of coarse and fine latents later adopted by multiscale tokenizers like AudioLM \cite{Borsos2023AudioLM:Generation}. Together, these works demonstrated that strong tokenizers can compress high-dimensional signals into compact representations suitable for large-scale generative modeling.

Most hierarchical tokenizers define hierarchy through fixed spatial mechanisms such as multiresolution grids, residual quantization, or decoder pyramids. HR-VQVAE \cite{Adiban2022HierarchicalGeneration}, RQ-VAE \cite{Lee2022AutoregressiveQuantization}, and VAR \cite{Tian2024VisualPrediction} progressively refine representations across scales and show that multiscale tokenization improves efficiency and inductive bias. However, these hierarchies remain fundamentally tied to Euclidean image structure rather than structured scientific domains whose meaningful units may be spatially distributed. A second line of work instead learns aggregation directly. Perceiver \cite{Jaegle2021Perceiver:Attention} compresses large inputs into latent bottlenecks via cross-attention, while DiffPool \cite{Ying2019HierarchicalPooling} learns soft hierarchical graph coarsening. These works suggest that tokenization need not rely on fixed partitions, but can instead emerge through learned many-to-few aggregation.

\subsection{Tokenization and Autoregressive Modeling for fMRI}

Recent fMRI foundation models primarily focus on pretraining objectives rather than on learning the tokenizer itself. BrainLM \cite{Caro2023BrainLM:Recordings} applies large-scale masked-prediction training to rs-fMRI and demonstrates transfer to clinical and demographic tasks, while Brain-JEPA \cite{Dong2024Brain-JEPA:Masking} introduces latent-target prediction with spatiotemporal masking. SLIM-Brain \cite{Wang2026SLIM-Brain:Analysis} moves toward atlas-free voxel-level modeling using selective hierarchical encoding. These works show that both atlas-based and voxel-level pretraining can transfer effectively, but generally treat representational units as fixed.

Several recent works move closer to explicit tokenization and compression. BrainCodec \cite{Nishimura2024BrainCodec:States} imports neural codec ideas into fMRI for denoising and reconstruction, while HST \cite{Yang2025HierarchicalQuantization} combines hierarchical state-space modeling with clustered VQ-VAE quantization for brain-state analysis. Earlier voxelwise VAE work \cite{Geenjaar2021VariationalSignatures} similarly demonstrated that learned latent rs-fMRI representations preserve information useful for downstream prediction. Complementary autoregressive approaches include FARM \cite{Garg2011Full-brainFMRI}, continuous-space autoregression \cite{Paugam2024ADataset}, voxel-level transformer prediction \cite{Sun2025Voxel-LevelTransformer}, and multimodal next-token modeling \cite{Yin2026BrainVista:Prediction}. Hierarchical functional organization has also been used as an inductive bias for supervised brain-network classification, e.g. DHGFormer \cite{Xue2026DHGFormer:Diagnosis} guides hierarchical graph modeling with Yeo-network structure and Zhu et al. \cite{Zhu2024Spatio-TemporalClassification} pool adaptively between ROI graphs. However, these operate on connectivity instead of voxel-level data. Together, these frameworks establish that compression and autoregressive modeling of fMRI signals are both feasible and useful.

\subsection{Recent Tokenization and Quantization Techniques}

Recent tokenizer research increasingly studies the geometry and scalability of the quantization bottleneck itself. LFQ/MAGVIT-v2 \cite{Yu2024LanguageGeneration}, FSQ \cite{Mentzer2023FiniteSimple}, BSQ \cite{Zhao2024ImageQuantization}, and spherical Leech quantization \cite{Zhao2025SphericalGeneration} demonstrate that bottleneck geometry strongly influences tokenizer stability, compression efficiency, and scalability. Other approaches focus on compactness and codebook utilization. TiTok \cite{Yu2024AnGeneration} compresses images into extremely short token sequences, VQGAN-LC \cite{Zhu2024Scaling99} scales to very large vocabularies, IBQ \cite{Shi2025ScalableQuantization} introduces differentiable code indices, and FQGAN \cite{Bai2024FactorizedGeneration} factorizes vocabularies into semantically enriched sub-codebooks. Recent work also increasingly co-designs tokenizers and autoregressive models. VAR \cite{Tian2024VisualPrediction}, for example, replaces raster next-token prediction with next-scale prediction over multiscale latent maps generated by a hierarchical VQ-VAE. However, these approaches remain largely tied to predefined image pyramids and grid-structured data, whereas rs-fMRI exhibits distributed low-SNR structure spanning voxel, parcel, and network scales.

\section{\modelname: Hierarchical Tokenization for fMRI}
\label{sec:framework}

\subsection{Problem Formulation}

We consider data defined over a structured domain represented as a graph
\(\mathcal{G} = (V, E)\), where \(V\) denotes a set of nodes and \(E\) encodes relationships between nodes. Each observation is a signal:
\[
X \in \mathbb{R}^{|V| \times d_{\text{in}}}
\]
where nodes \(i \in V\) are associated with feature vectors \(X_i\). In rs-fMRI, nodes correspond to voxels, and edges reflect spatial adjacency or implicit functional relationships. Our goal is to construct a \emph{hierarchical tokenization} of \(X\), producing a sequence of progressively coarser representations:
\[
Z^{(0)} \rightarrow Z^{(1)} \rightarrow \cdots \rightarrow Z^{(L)}
\]
where \(Z^{(0)} = X\), and \(Z^{(l)} \in \mathbb{R}^{N_l \times d}\) denotes a set of \(N_l\) tokens at level \(l > 0\), with \(N_0 > N_1 > \cdots > N_L\). \camera{In our implementation for fMRI, the graph notation describes grouping structure, and attention within each level is global rather than restricted by an explicit adjacency matrix.}

\subsection{Hierarchical Tokenization as Aggregation}

We define a general tokenization operator at level \(l\) as $Z^{(l+1)} = \mathcal{T}^{(l)}(Z^{(l)}, \mathcal{G})$ which maps fine-scale representations to coarser tokens.  We consider two complementary regimes:

\paragraph{Fixed (prior-based) tokenization.}
Tokenization may be defined by a partition of nodes into groups \(\{S_k \subset V\}_{k=1}^{N_{l+1}}\), yielding:
\[
Z^{(l+1)}_k = \mathrm{Pool}\bigl(\{Z^{(l)}_i : i \in S_k\}\bigr)
\]
where \(\mathrm{Pool}\) is an aggregation operator (e.g., mean or attention pooling). In structured domains such as the brain, these partitions may be derived from large-scale datasets (e.g., functional atlases), providing a strong prior over meaningful organization. A large body of resting-state fMRI literature focuses on deriving large scale groupings of brain regions at various scales: as small as seven networks \cite{ThomasYeo2011TheConnectivity} and as large as 1024 \cite{Dadi2020Fine-grainAnalysis} or even 3000 parcels \cite{Blumensath2013SpatiallyFMRI}. Similarly, we instantiate our framework for resting-state fMRI using a three-level hierarchy spanning voxel, parcel, and network scales. For each subject, we assume a preprocessed 4D resting-state fMRI volume 
\[
X \in \mathbb{R}^{X \times Y \times Z \times T}
\]
in standard space (e.g., 2\,mm MNI), along with a gray-matter mask \(M \in \{0,1\}^{X \times Y \times Z}\). Only voxels with \(M=1\) are retained, yielding \camera{\(N_V=|V|\)} voxels per subject. In our experiments, \camera{the mask retains \(N_V=190{,}468\)} voxels and \camera{scans have \(T \in [922,1200]\)} timepoints.

% We then construct representations at three scales as previously described:

% \paragraph{Voxel scale.}
% \[
% X^{(v)} \in \mathbb{R}^{T \times V}
% \]

% \paragraph{Parcel scale.}
% \[
% X^{(p)}_{t,p} = \frac{1}{|\mathcal{V}_p|} \sum_{v:\,p(v)=p} \tilde{x}_t(v),
% \quad X^{(p)} \in \mathbb{R}^{T \times 1024}
% \]

% \paragraph{Network scale.}
% \[
% X^{(n)}_{t,k} = \frac{1}{|\mathcal{P}_k|} \sum_{p:\,n(p)=k} X^{(p)}_{t,p},
% \quad X^{(n)} \in \mathbb{R}^{T \times 17}
% \]

\paragraph{Learned tokenization via cross-attention.}
Alternatively, tokenization can be defined as a learned aggregation:
\[
Z^{(l+1)} = \mathrm{Attn}(Q^{(l)}, Z^{(l)}),
\]
where \(Q^{(l)} \in \mathbb{R}^{N_{l+1} \times d}\) are learned query embeddings, and \(\mathrm{Attn}\) denotes multi-head cross-attention. Each token aggregates information from all tokens at level \(l\) with learned, input-dependent weights, allowing the model to learn distributed and non-local structure directly from data.

In our architecture, at the bottom level, voxel-level sub-parcel tokens are first encoded using a temporal encoder operating on local voxel groups over short temporal windows. The resulting token representations are then aggregated hierarchically to produce parcel- and network-level representations. Finally, we also consider a soft hierarchy setting, in which the same learned cross-attention architecture is used, but representations are regularized toward atlas-derived groupings. 

Overall, these two regimes represent complementary design choices: fixed tokenization leverages strong domain priors; learned tokenization can capture distributed, data-driven structure. Both are instances of one hierarchical aggregation operator, which generalizes the fixed local neighborhoods of CNNs and the predefined neighborhoods of graph networks to aggregation that may be fixed or learned at each scale. This flexibility is especially important in domains such as brain data, where meaningful structure is often distributed and not strictly localized in space.

\subsection{Multiscale Representation}

Each level of the hierarchy induces an implicit coarsening of the underlying graph. Tokens at level \(l+1\) can be interpreted as nodes in a coarsened graph \(\mathcal{G}^{(l+1)}\), where each coarse node represents a subset or distribution over nodes at level \(l\). Unlike classical graph coarsening methods, which rely on fixed clustering heuristics, our formulation allows tokenization to be defined either through explicit priors or learned aggregation, enabling flexible multiscale representations. 

\paragraph{Voxel-level}
At the lowest level, we construct voxel-level tokens by partitioning each parcel into fixed-size voxel groups of size \(K=16\), preserving fine-grained spatial information that is typically lost in standard parcellations that average much larger groups of voxels. For each parcel, voxel coordinates are normalized and ordered deterministically, after which they are divided into contiguous groups, padding the final group if necessary. This results in \camera{\(N_C=12{,}358\) spatial chunks, with 7,260 padded slots. The number of temporal codes} per \camera{chunk depends on the operating setting (Section~\ref{sec:windowing}). These ordered voxel lists need not form compact 3D cubes.} 

\paragraph{Parcel-level}
At the intermediate level, we use DiFuMo-1024 parcels \cite{Dadi2020Fine-grainAnalysis}.

\paragraph{Network-level}
At the top level, we use large-scale functional networks, specifically the Yeo-17 \camera{labels supplied with DiFuMo} \cite{ThomasYeo2011TheConnectivity}, \camera{plus an extra group for unassigned parcels, giving 18 groups in total}. Importantly, this hierarchy is anchored in well-established functional atlases, which provide a strong prior over brain organization.

\camera{With $C,P,N$ denoting chunks, parcels, and networks, the fixed encoder applies a temporal CNN/chunk projection $E$, factorized transformers $F_\ell$, and feature means $A_P,A_N$:

\begin{equation}
z_C=F_C(E(x)),\qquad z_P=F_P(A_Pz_C),\qquad z_N=F_N(A_Nz_P).
\label{eq:encoder}
\end{equation}

The features have width 192, and transformer depths are 3/2/1 with each depth comprising a spatial and a temporal attention block. Learned pools replace the means with cross-attention. Appendix~\ref{app:architecture} specifies all blocks and positions.}

\paragraph{Discretizations at multiple levels}
Vector \cite{Oord2018NeuralLearning}, binary spherical \cite{Zhao2024ImageQuantization}, or other quantization \cite{Zhao2025SphericalGeneration} can be applied at intermediate levels to obtain discrete tokens\camera{; our hierarchical models use BSQ at all three levels}. In practice, BSQ discretizes representations by projecting features onto a binary hypersphere using a straight-through sign operator, producing compact discrete tokens while maintaining stable optimization and efficient codebook utilization. Compared to conventional approaches, BSQ enables substantially higher compression without codebook collapse, making it especially attractive for extremely high-dimensional data such as fMRI.

% To ensure that tokenization preserves information about the original signal, we learn a hierarchical decoder:
% \[
% \hat{Z}^{(l)} = \mathcal{D}^{(l)}(Z^{(l+1)}), \quad \hat{X} = \mathcal{D}(Z^{(L)}),
% \]
% and optimize a reconstruction objective:
% \[
% \mathcal{L}_{\text{recon}} = \|X - \hat{X}\|_2^2.
% \]

\subsection{Optimization and Regularization}

Decoding proceeds hierarchically across scales: network-level predictions are first reconstructed from top-level tokens, parcel-level predictions are modeled as residuals relative to broadcast network predictions, and voxel-level reconstructions are modeled as residuals relative to parcel-level signals. This residual formulation encourages coarse representations to capture low-frequency global structure, while finer scales model higher-frequency deviations, \camera{and defines a coarse-to-fine residual decomposition. Every reconstruction path passes through discrete codes (Appendix~\ref{app:architecture})}, with loss:

\begin{equation}
  \camera{\mathcal L_{\mathrm{total}}
  = \mathcal L_{\mathrm{recon}} + \mathcal L_{\Delta}
  + \lambda_{\mathrm{vq}}\mathcal L_{\mathrm{vq}}
  + \sigma\mathcal L_{\mathrm{align}}
  + \lambda_{\mathrm{mid}}\mathcal L_{\mathrm{mid}}}
  \label{eq:loss}
\end{equation}

\paragraph{Reconstruction terms}
Mean-squared- or mean-absolute-error (MSE/MAE) supervision individually at voxel, chunk, parcel, and network scales encourages the hierarchy to preserve both fine-grained and large-scale functional organization. The temporal consistency term $\mathcal{L}_\Delta$ encourages preserving local temporal dynamics in noisy voxel signals. Let $V,C,P,N$ denote the voxel, chunk, parcel, and network reconstruction terms. Then, the terms, with weights in Appendix~\ref{app:training}, Table~\ref{tab:training}, are defined:

\begin{align}
\mathcal{L}_{\mathrm{recon}} &= \sum_{\ell \in \{V,C,P,N\}}\lambda_\ell
\mathcal{L}_{\ell}\!\left(\hat{\mathbf{X}}_{(\ell)}, \mathbf{X}_{(\ell)}\right) \\
\mathcal{L}_{\Delta} &= \sum_{\ell \in \{V,C\}} \lambda_{\Delta,\ell}\mathcal{L}_{\ell}\!\left(\Delta_t \hat{\mathbf{X}}_{\ell}, \Delta_t \mathbf{X}_{\ell}\right)
\end{align}

\paragraph{Soft cross-scale alignment}
In the soft hierarchy regime, atlas-derived representations are treated as \camera{stopped-gradient} targets and do not receive gradients. The alignment objective therefore acts as a soft structural prior, encouraging learned \camera{representations} to remain close to known functional organization while still permitting flexible reassignment through attention-based pooling. In contrast, the fully learned hierarchy omits this term and learns cross-scale organization purely from reconstruction objectives, without the atlas alignment penalty.
\begin{align}
\mathcal{L}_{\mathrm{align}}
=\;
\left\lVert
\mathbf{Z}_{\mathrm{parcel}}^{\mathrm{learned}}
-\mathbf{Z}_{\mathrm{parcel}}^{\mathrm{atlas}}
\right\rVert_2^2
+\left\lVert
\mathbf{Z}_{\mathrm{network}}^{\mathrm{learned}}
-\mathbf{Z}_{\mathrm{network}}^{\mathrm{atlas}}
\right\rVert_2^2
\end{align}

\paragraph{BSQ objective}
The quantization term is summed across all levels:
\begin{align}
\mathcal{L}_{\mathrm{vq}}
=
\sum_{\ell \in \{\mathrm{voxel,parcel,network}\}} \mathcal{L}_{\mathrm{BSQ}}^{(\ell)}.
\end{align}

For each level \(\ell\), with $\hat{\mathbf{z}}^{(\ell)} = \mathrm{sign}\!\left(\mathbf{z}^{(\ell)}\right)$, namely $\mathcal{L}_{\mathrm{BSQ}}^{(\ell)}$ is a weighted average of the commitment loss and the entropy penalty, $\mathcal{L}_{\mathrm{BSQ}}^{(\ell)} =
\mathcal{L}_{\mathrm{commit}}^{(\ell)}
+ \frac{\zeta^{(\ell)}}{\tau_{\mathrm{inv}}^{(\ell)}}\mathcal{L}_{\mathrm{entropy}}^{(\ell)}$:

\begin{align}
\mathcal{L}_{\mathrm{commit}}^{(\ell)}
= \beta^{(\ell)}
\left\lVert
\mathrm{sg}\!\left(\hat{\mathbf{z}}_{\mathrm{scaled}}^{(\ell)}\right)
- \mathbf{z}^{(\ell)}
\right\rVert_2^2, \quad
\mathcal{L}_{\mathrm{entropy}}^{(\ell)}
=
\gamma_0^{(\ell)} H_{\mathrm{persample}}^{(\ell)}
- \gamma^{(\ell)} H_{\mathrm{codebook}}^{(\ell)}
\end{align}

where \(\mathrm{sg}(\cdot)\) is stop-gradient and $\beta$, $\gamma_0$, $\gamma$ and $\zeta$ are hyperparameters. BSQ quantization is applied independently at the voxel-, parcel-, and network-token levels prior to hierarchical decoding (Figure~\ref{fig:intro}).

\section{Experiments}
\label{sec:experiments}

\subsection{Data and Preprocessing}

We use $N=1096$ resting-state fMRI BOLD recordings from the Human Connectome Project Young-Adults (HCP-YA) \cite{VanEssen2013TheOverview}\camera{, with one run per subject and TR $0.72$\,s}. We \camera{use 877/110/109 subjects} for training\camera{/validation/test}, and train for 10 epochs with a \camera{$2\times10^{-4}$} learning rate. The datasets are partitioned at the subject level to prevent any repeated or longitudinal recordings from the same individual from appearing in both training and testing sets. All datasets are preprocessed using standardized pipelines \cite{Abbasi2025FMRIControl,Esteban2019FMRIPrep:MRI}, and all models and baselines are trained on identical data splits using approximately matched compression ratios to ensure fair comparison across tokenizer families (see Appendix~\ref{app:bits}). Additional details are presented in Appendix~\ref{app:data}.

\subsection{Baselines}

We compare against flat tokenizer baselines at each storage budget and with four atlas-derived representations for linear probing.

\camera{\textbf{Rate-matched flat tokenizers.}}
\camera{Both operate directly on voxel chunks. Flat BSQ uses 19 sign bits per position; flat RVQ uses two residual codebooks with 1,024 and 512 entries (10+9 bits). Flat BSQ uses eight temporal positions per chunk at $52\times$ and one at $415\times$. The bit budgets are matched to within 0.2\%; model size and training objectives can differ.}

\noindent{\bf{MS-VQVAE.}}
A hierarchical VQ-VAE baseline inspired by multiscale autoregressive image models such as VAR \cite{Tian2024VisualPrediction}. This baseline progressively compresses representations across scales using learned hierarchical latent maps, but does not incorporate structured neurobiological priors or explicit atlas-based organization. To provide a bitrate-matched multiscale VQ baseline, we implement a two-level hierarchical VQ-VAE following the progressive latent compression strategy used in VAR \cite{Tian2024VisualPrediction}. The model has two discrete scales: a middle scale with one token per voxel chunk (12,358 per window), quantized with a two-stage residual VQ (2×512 codes, 18 bits), and a top scale with 1,024 tokens obtained by learned cross-attention pooling over the chunk tokens, quantized with a 4,096-code VQ (12 bits). Both quantizers use EMA codebook updates with dead-code restarts on a 32-dimensional projection, giving $\approx 415\times$ compression. Unlike our hierarchical BSQ framework, which uses three explicit scales (voxel, parcel, and network), the MS-VQVAE baseline uses only two discrete latent scales and does not incorporate structured priors or cross-scale alignment objectives.

\noindent{\bf{DiFuMo Baselines.}}
\camera{We use Yeo network means, DiFuMo parcel means, and two training-fitted PCA representations derived from parcel time series or parcel functional connectivity.}

\subsection{Temporal Windowing}
\label{sec:windowing}

Training is performed on overlapping temporal windows of length $W=32$ with stride $s=16$. For a sequence of length $T$, this yields $N_{\mathrm{win}} = \lfloor \frac{T-W}{s} \rfloor + 1$ training windows per scan, corresponding to approximately \camera{56}--74 windows in our setting. Each training sample consists of voxel-level windows, sub-parcel token groups, and corresponding parcel- and network-level summaries. Within the encoder, temporal windows are partitioned into smaller segments that are summarized into a fixed number of latent temporal tokens at each scale.

\camera{HBAR-52 stores eight temporal codes per spatial unit (one per four TRs), with 18, 12, and 6 bits at the chunk, parcel, and network levels. HBAR-415 stores one code per 32-TR window, with 18/12/24 bits. The compact setting also changes temporal pooling, the parcel decoder, and residual supervision, and is therefore a separately trained system rather than a pure rate ablation (Appendix~\ref{app:architecture}).}

\subsection{\camera{Evaluation and uncertainty}}

\camera{Reconstruction uses the same 16 windows for each of 109 test subjects. All models are evaluated by pooling their final voxel reconstructions with the true atlas: voxels are equally weighted within parcels and parcels are equally weighted within networks. Pearson correlation flattens time and units per window; scores are averaged over windows within each subject and then across subjects. We distinguish three-seed sample SD from per-model subject-bootstrap CIs and paired contrasts (Appendix~\ref{app:metrics}).}

\camera{Linear probes use mean-pooled post-quantization features and logistic regression, with training-only standardization and five-fold inner-CV regularization selection. We report AUROC for sex and for three-bin age for seed 0. The AR readout applies the same type of probe to mean-pooled hidden states of a causal model trained on bottom-level tokens over four consecutive 32-TR windows. Probes and atlas baselines use matched stride-64 windows; full settings are in Appendix~\ref{app:probes}.}

% \subsection{Implementation Details}

% Our default configuration uses \(P=1024\) parcels, \(K=16\) voxels per sub-parcel token, 17 networks, window length \(W=32\), stride \(s=16\), and batch size \(B=2\).

\vspace{2em}
\begin{table*}[h!]
\centering
\small
\setlength{\tabcolsep}{4pt}

\begin{tabular*}{\textwidth}{@{\extracolsep{\fill}}lcccc@{}}
  \toprule
  Model & Ratio & Voxel Corr$\uparrow$ & Parcel Corr$\uparrow$ & Net Corr$\uparrow$ \\
  \midrule
  Window-Mean Oracle & --- & $0.503$ & $0.578$ & $0.522$ \\
  Random Baseline & --- & $0.000$ & $0.000$ & $-0.002$ \\
  \midrule
  HBAR-415 (fixed)  & $415\times$ & $0.405\pm0.004$ & $\mathbf{0.746\pm0.003}$ & $\mathbf{0.895\pm0.002}$ \\
  Flat BSQ          & $415\times$ & $0.394\pm0.002$ & $0.627\pm0.005$ & $0.710\pm0.010$ \\
  Flat RVQ          & $415\times$ & $\underline{0.413\pm0.009}$ & $\underline{0.691\pm0.002}$ & $\underline{0.803\pm0.010}$ \\
  MS-VQVAE          & $415\times$ & $\mathbf{0.427\pm0.001}$ & $0.594\pm0.000$ & $0.641\pm0.004$ \\
  \midrule
  HBAR-52 (fixed)   & $52\times$  & $\underline{0.526\pm0.004}$ & $\mathbf{0.920\pm0.003}$ & $\underline{0.965\pm0.005}$ \\
  Learned pooling   & $52\times$  & $0.519\pm0.005$ & $0.876\pm0.005$ & $0.938\pm0.009$ \\
  Soft pooling      & $52\times$  & $0.487\pm0.063$ & $0.796\pm0.022$ & $0.898\pm0.030$ \\
  Shuffled grouping & $52\times$  & $0.523\pm0.006$ & $0.886\pm0.004$ & $\underline{0.966\pm0.011}$ \\
  Flat BSQ          & $52\times$  & $\mathbf{0.529\pm0.002}$ & $\underline{0.890\pm0.002}$ & $\mathbf{0.968\pm0.010}$ \\
  \bottomrule
\end{tabular*}

\caption{Reconstruction performance across tokenizers and baselines on 109 held-out subjects with 16 windows per subject (1,744 windows). Higher correlation is better. Tokenizer results report mean $\pm$ sample standard deviation across three training seeds; ratios denote rounded per-window code-payload compression. The first two rows are target-informed, model-free references: the window-mean oracle predicts each unit's true within-window temporal mean, while the random reference samples Gaussian noise matched to the target window's global mean and standard deviation at each scope. Reference rows report point estimates. Bold and underlining indicate the highest and second-highest mean within each tokenizer rate block, respectively, not statistical significance.
}
\label{tab:recon}
\end{table*}

\begin{table*}[t]
\centering
\small
\setlength{\tabcolsep}{6pt}

\begin{tabular}{lcccc}
\toprule
& \multicolumn{2}{c}{Linear Probing}
& \multicolumn{2}{c}{AR Readout} \\
\cmidrule(lr){2-3}\cmidrule(lr){4-5}
Model
& Age AUROC$\uparrow$
& Sex AUROC$\uparrow$
& Age AUROC$\uparrow$
& Sex AUROC$\uparrow$ \\
\midrule
Yeo-Networks
& \scoreci{0.530}{0.44}{0.61}
& \scoreci{0.551}{0.44}{0.66}
& --- & --- \\
DiFuMo
& \scoreci{0.561}{0.47}{0.65}
& \scoreci{0.559}{0.45}{0.67}
& --- & --- \\
DiFuMo $\rightarrow$ PCA
& \scoreci{0.580}{0.49}{0.67}
& \scoreci{0.574}{0.47}{0.68}
& --- & --- \\
Parcel-FC + PCA
& \scoreci{0.683}{0.61}{0.75}
& \scoreci{\mathbf{0.956}}{0.92}{0.99}
& --- & --- \\

\midrule

HBAR-415 (fixed)
& \scoreci{0.482}{0.41}{0.56}
& \scoreci{\underline{0.701}}{0.60}{0.79}
& \scoreci{\mathbf{0.584}}{0.50}{0.66}
& \scoreci{\mathbf{0.638}}{0.53}{0.74} \\

Flat BSQ-415
& \scoreci{\mathbf{0.607}}{0.53}{0.68}
& \scoreci{0.648}{0.54}{0.75}
& \scoreci{0.506}{0.43}{0.59}
& \scoreci{0.556}{0.45}{0.66} \\

Flat RVQ-415
& \scoreci{\underline{0.557}}{0.47}{0.64}
& \scoreci{\mathbf{0.722}}{0.62}{0.81}
& \scoreci{0.503}{0.42}{0.58}
& \scoreci{\underline{0.636}}{0.53}{0.74} \\

MS-VQVAE-415
& \scoreci{0.540}{0.45}{0.63}
& \scoreci{0.605}{0.50}{0.71}
& \scoreci{\underline{0.515}}{0.43}{0.60}
& \scoreci{0.545}{0.43}{0.66} \\

\midrule

HBAR-52 (fixed)
& \scoreci{\underline{0.606}}{0.52}{0.68}
& \scoreci{\mathbf{0.795}}{0.71}{0.87}
& \scoreci{\underline{0.557}}{0.48}{0.63}
& \scoreci{\mathbf{0.792}}{0.70}{0.87} \\

Learned pooling ($52\times$)
& \scoreci{0.593}{0.51}{0.68}
& \scoreci{0.746}{0.65}{0.83}
& \scoreci{\mathbf{0.588}}{0.50}{0.67}
& \scoreci{0.742}{0.65}{0.83} \\

Soft pooling ($52\times$)
& \scoreci{\mathbf{0.655}}{0.56}{0.75}
& \scoreci{0.744}{0.65}{0.83}
& \scoreci{0.507}{0.43}{0.59}
& \scoreci{0.711}{0.61}{0.81} \\

Shuffled grouping ($52\times$)
& \scoreci{0.588}{0.50}{0.67}
& \scoreci{\underline{0.769}}{0.68}{0.85}
& \scoreci{0.525}{0.43}{0.61}
& \scoreci{\underline{0.752}}{0.66}{0.84} \\

Flat BSQ-52
& \scoreci{0.580}{0.50}{0.66}
& \scoreci{0.737}{0.64}{0.82}
& \scoreci{\underline{0.557}}{0.47}{0.64}
& \scoreci{0.640}{0.53}{0.74} \\

\bottomrule
\end{tabular}

\caption{Demographic prediction performance on 109 held-out test subjects using linear probing and autoregressive (AR) readout. Age is a three-bin classification task; values are AUROC with 95\% subject-bootstrap confidence intervals. Linear probes use pooled tokenizer features or the indicated atlas-based representations. AR readouts use mean-pooled final-layer hidden states from causal sequence models trained on bottom-level token sequences. Tokenizer results use seed 0 in both panels. Bold and underlining indicate the highest and second-highest point estimates within each tokenizer rate block, respectively.}
\label{tab:downstream}
\end{table*}

\section{Results}
\label{sec:results}

\subsection{Reconstruction}

\camera{At approximately $415\times$ compression, HBAR achieves parcel and network correlations of 0.746 and 0.895, compared with 0.691 and 0.803 for flat RVQ and 0.627 and 0.710 for flat BSQ (Table~\ref{tab:recon}). Voxel correlation is slightly lower than for RVQ (0.405 versus 0.413) and higher than for flat BSQ (0.394), with the MS-VQVAE baseline achieving the highest correlation (0.427).} At this bit budget, HBAR’s gain over the baselines is concentrated at the coarser scales. \camera{At $52\times$, flat BSQ gives similar voxel and network correlations to HBAR, while fixed HBAR retains the highest parcel correlation (0.920).}

At 52×, shuffled grouping yields nearly the same voxel correlation as fixed atlas grouping (0.523 versus 0.526) and nearly the same network correlation (0.966 versus 0.965). The difference appears at the parcel scale: preserving the anatomical chunk-to-parcel assignments raises correlation to 0.920. A spectral analysis is included in Appendix~\ref{app:spectral}.

\subsection{Downstream Performance}

We next evaluate whether the learned representations support downstream modeling beyond reconstruction. We consider two evaluation settings: (\emph{i}) linear probing using mean-pooled token embeddings, and (\emph{ii}) autoregressive modeling using a causal sequence model trained directly on the tokenized representations. \camera{Table~\ref{tab:downstream} reports demographic classification from both readouts.} 

The relative ordering of models differs between linear probing and autoregressive evaluation. \camera{At $52\times$, fixed HBAR achieves strong sex AUROC under both linear probing (0.795) and AR readout (0.792), leading the methods reported in Table~\ref{tab:downstream}. At $415\times$, HBAR achieves the highest observed AR AUROC for both age (0.584) and sex (0.638), despite not leading under linear probing. These results suggest that the utility of compressed representations depends on the downstream readout, and that linear decodability alone may not fully capture their suitability for sequence modeling. Parcel-FC with PCA remains the strongest overall linear baseline for sex prediction (AUROC 0.956), consistent with prior work demonstrating the predictive value of resting-state connectivity \cite{Weis2020SexConnectivity,He2020DeepDemographics}.}

Performance on age prediction is comparatively modest across all methods. At $415\times$, HBAR has the highest observed AR AUROC (0.584), while flat BSQ leads under linear probing (0.607). At $52\times$, soft pooling leads under linear probing (0.655) and learned pooling leads with the AR readout (0.588). HCP-YA’s narrow, binned age range limits this task (see Limitations).

\begin{figure}[t]
  \centering
  \includegraphics[width=\linewidth]{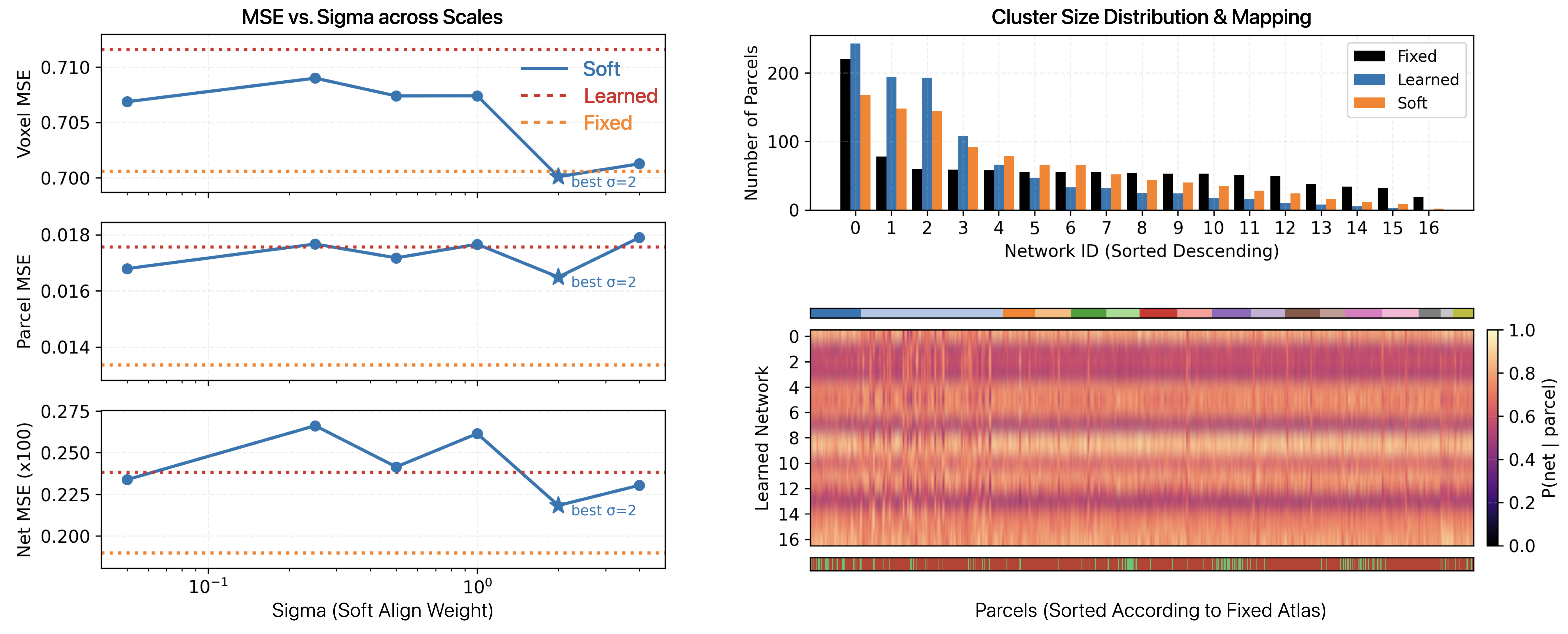}
  \caption{
  \textbf{Comparison of fixed, learned, and softly aligned hierarchical tokenization strategies.}
  (\textbf{Left}) Reconstruction error across voxel, parcel, and network scales as a function of the soft-alignment weight $\sigma$. Larger $\sigma$ imposes stronger alignment to the atlas prior. Across the sweep, $\sigma=2$ gives the lowest soft-hierarchy error at all three scales; the soft hierarchy matches the fixed hierarchy at voxel level but remains worse at parcel and network scales for every $\sigma$.
  (\textbf{Top right}) Distribution of parcel assignments across learned network slots for fixed, learned, and soft hierarchies. Because learned network indices are permutation-invariant, we compare distributions rather than direct one-to-one correspondence. Both learned and soft hierarchies exhibit skewed assignment patterns similar to the atlas-based hierarchy, suggesting convergence toward structured large-scale organization.
  (\textbf{Bottom right}) Soft parcel-to-network assignment matrix after Hungarian alignment between learned and atlas-derived network indices. Parcels are sorted according to fixed atlas labels, and colors indicate the probability of assignment to each aligned learned network. The diffuse structure and sparse hard matches indicate that fully learned hierarchies only partially recover atlas-like organization, despite exhibiting similar global assignment statistics.
  }
  \label{fig:hierarchy}
\end{figure}

\subsection{Learned vs. Fixed Hierarchies}

To better understand the role of structured priors, we compare fully learned, fixed atlas-based, and softly aligned hierarchical tokenization strategies (Fig.~\ref{fig:hierarchy}). The soft hierarchy introduces a tunable alignment parameter $\sigma$ controlling the tradeoff between learned flexibility and adherence to atlas structure. As shown in Fig.~\ref{fig:hierarchy} (left), varying $\sigma$ changes reconstruction only modestly and non-monotonically, and no setting closes the gap to the fixed hierarchy at parcel or network scale. The best-performing configurations at coarser scales use predefined atlas organization, reinforcing the importance of structured priors in modeling large-scale brain activity.

Because learned network indices are permutation-invariant, direct comparison between learned and atlas-derived hierarchies is not meaningful. Instead, we analyze global assignment structure and align learned network slots to atlas networks using Hungarian matching \cite{Kuhn1955TheProblem}. Interestingly, the learned hierarchy exhibits assignment distributions broadly similar to the fixed atlas hierarchy, with several dominant network slots containing disproportionately large mass (Fig.~\ref{fig:hierarchy}, top right). This suggests that the model partially recovers large-scale organizational structure even without explicit supervision. However, the aligned soft assignment matrix reveals that learned parcel-to-network mappings remain substantially more diffuse than atlas-derived organization (Fig.~\ref{fig:hierarchy}, bottom right). Rather than recovering localized atlas-like partitions, learned hierarchies distribute mass, indicating that reconstruction objectives alone are insufficient to recover canonical functional organization.

\subsection{Model Efficiency}

\modelname contains 5.6M/7.3M parameters for the $52\times$/$415\times$ models, and processes a temporal window in ~231\,ms on 1x H100 GPU in our implementation. This corresponds to approximately 9 seconds to tokenize an entire 1200-frame resting-state scan, making the approach practical for patient-level analysis and large-scale neuroimaging datasets alike while operating directly on voxel-level signals.

\section{Discussion}
\label{sec:discussion}

We introduced a hierarchical tokenization for structured data and instantiated it for resting-state fMRI through a voxel-to-parcel-to-network hierarchy. \camera{At $415\times$ compression, the hierarchy preserved parcel and network signals more accurately than baselines. At $52\times$, flat BSQ was competitive, although HBAR still maintained the best parcel reconstruction. At both scales, HBAR maintained strong downstream performance. The preferred tokenizer therefore depends not only on compression rate and reconstruction scale, but also on how the resulting codes are used.} In summary, we demonstrated that compact discrete representations can retain information from voxel-level fMRI while remaining practical for large-scale modeling.

A second key finding is that neurobiological priors can improve structural fidelity while retaining downstream predictive utility. At $52\times$ compression, the fixed anatomical hierarchy reconstructed parcel signals more accurately than learned or soft alternatives. Shuffling chunk-to-parcel assignments reduced parcel correlation across all three seeds, indicating that the benefit extends beyond the grouping sizes alone. Importantly, this structural advantage does not come at the expense of demographic information: fixed HBAR achieves strong sex prediction under both linear and autoregressive readouts, while at $415\times$ it leads the AR comparisons on both tasks. Together, these findings suggest that anatomical structure provides a useful inductive bias for compact representation learning.

More broadly, these findings highlight a useful design principle for representation learning in scientific domains. Unlike images or videos, where multiscale structure is often defined by spatial resolution, fMRI has meaningful organization across distributed functional systems. Our results indicate that hierarchical tokenization should not simply import grid-based multiscale designs, but should instead exploit domain structure. In this sense, atlas-based hierarchies are not merely preprocessing choices; they may provide powerful inductive biases that can guide interpretable and effective token formation.

\paragraph{Limitations}
This study focuses on resting-state fMRI from HCP-YA, which has a relatively narrow young-adult age range and discretized age labels. As a result, age prediction is a challenging and limited downstream task. Our experiments also use a fixed atlas hierarchy based on DiFuMo-1024 and Yeo-17, and although these are well-established priors, other parcellations or subject-specific functional networks may yield different tradeoffs. Additionally, our scale comparison contrasts one-scale flat tokenizers with the three-scale hierarchy and is not a sweep over hierarchy depth. Finally, while we evaluate autoregressive modeling and demographic prediction, we do not yet test broader clinical, cognitive, or cross-dataset transfer settings where the value of voxel-preserving tokenization may be even larger.

\paragraph{Future work}
Future work should evaluate hierarchical tokenization across more diverse neuroimaging cohorts, including lifespan, aging, and disease datasets. The framework also does not depend on volumetric input: because it assumes only a graph, the nodes could equally be CIFTI grayordinates with a surface-based parcellation at the intermediate level, which would likely improve cross-subject cortical correspondence at the cost of the subcortical and cerebellar coverage our gray-matter mask retains. A particularly important direction is subject-adaptive tokenization, where atlas priors provide initialization or regularization while allowing individualized deviations in functional organization. Another promising direction is to couple the tokenizer more tightly to downstream autoregressive objectives, so that the hierarchy directly shapes temporal brain dynamics modeling. Finally, while this paper focuses on fMRI, the framework is naturally applicable to other structured scientific signals, including electrophysiology, spatial transcriptomics, and sensor networks, where meaningful units are distributed and multiscale.

\paragraph{Conclusion}
Hierarchical tokenization provides a compact and expressive way to represent high-dimensional voxel-level fMRI. By preserving sub-parcel information while organizing representations across parcel and network scales, HBAR preserves meaningful brain signals under substantial compression while retaining information useful for downstream demographic prediction and autoregressive modeling. At the same time, our comparison of fixed, learned, and soft hierarchies shows that strong domain priors remain critical: established functional atlases provide structure that is difficult to recover from reconstruction alone. These results argue for a hybrid view of scientific representation learning, where learned tokenizers are most effective when paired with principled domain structure.

\newpage
\paragraph{Acknowledgments.}
This work was supported by the Stanford Institute for Human-Centered AI (HAI) Hoffman-Yee Award, UST, and National Institutes of Health (NIH) Grant AG089169. KS was supported by the National Science Foundation (NSF) Graduate Research Fellowship Program under Grant No. DGE-2146755. Any opinions, findings, and conclusions or recommendations expressed in this material are those of the author(s) and do not necessarily reflect the views of the NSF or NIH.

% REFERENCES

\bibliography{references}
\bibliographystyle{plainnat}

%%%%%%%%%%%%%%%%%%%%%%%%%%%%%%%%%%%%%%%%%%%%%%%%%%%%%%%%%%%%

\appendix

\section{\camera{Data and spatial layout}}
\label{app:data}

\subsection{\camera{Cohort and preprocessing}}
\camera{We use one selected left-to-right resting-state run per HCP-YA subject (the longest eligible run in the data), yielding 1,096 subjects. Of these, 1,091 runs have 1,200 TRs and five have 922--1,124 TRs, for 1,314,326 TRs in total with a TR of 0.72\,s. This produces 877 training, 110 validation, and 109 test subjects. With 32-TR windows, stride 16, and no incomplete trailing windows, the data comprises 64,855/8,125/8,066 train/validation/test windows (81,046 total). Splits are by subject, not by window. Family identifiers were not available to enforce family-disjoint splits.}

\camera{Beyond fMRIPrep, HBAR preprocessing retains every voxel in the gray-matter mask, z-scores each voxel over time within its scan with $\epsilon=10^{-6}$, and stores float16 arrays. HBAR adds no denoising, filtering, detrending, or censoring. Normalization uses the complete individual scan, including time points outside a given window. The input grid is MNI-152 at 2\,mm, $97\times115\times97$ (1,082,035 grid positions), of which the template mask retains 190,468 unique voxels. The mask derives from a template gray-matter mask resampled with nearest-neighbor interpolation, and the retained volume includes cortical, cerebellar, and subcortical regions; a small number of voxels in white-matter/CSF may also remain near mask boundaries.}

\subsection{\camera{Atlas construction and padding}}
\camera{The DiFuMo-1024 continuous maps are cubically resampled to the BOLD grid, then converted to a hard, nonoverlapping partition by an argmax across maps. Within each parcel, voxel coordinates are min--max normalized along each axis, ordered by $x+10^{-2}y+10^{-4}z$, and divided into consecutive groups of 16. Chunks are ordered voxel lists, not necessarily compact 3D cubes or a space-filling curve. Parcel order and slot order are deterministic, and each voxel belongs to exactly one chunk.}

\camera{There are 12,358 chunks with 197,728 slots, comprising 190,468 valid voxels and 7,260 zero-filled padding slots. Of the chunks, 925 are partial and no chunk is entirely padding. Fourteen of the 1,024 parcels have no retained voxels, and their pooled signal is defined to be zero. They remain as token slots and are not attention-masked, so their codes still count toward storage. The model's CNN processes zero padding without an encoder mask; voxel reconstruction drops invalid slots, whereas chunk-level training losses include them.}

\camera{The network mapping uses 17 Yeo networks plus an eighteenth group for unassigned parcels. That group contains 220 parcels, including all 14 empty parcels, and is not interpreted as a physiological network. For a nonempty parcel $p$ with voxel set $\mathcal V_p$, the target signal is:

\begin{equation}
y_P[p,t]=\frac{1}{|\mathcal V_p|}\sum_{v\in\mathcal V_p}x[v,t],\qquad
y_N[n,t]=\frac{1}{|\mathcal P_n|}\sum_{p\in\mathcal P_n}y_P[p,t]
\label{eq:truepool}
\end{equation}

The network mean is unweighted over parcels, not over voxels. The unassigned group's target includes its 14 zero parcel signals in its denominator of 220. Feature pooling in Eq.~\eqref{eq:encoder}, by contrast, is an unweighted mean of child tokens; it does not weight partial chunks by voxel count.}

\section{\camera{Architecture and decoding details}}
\label{app:architecture}

\subsection{\camera{\texorpdfstring{The 52$\times$ configuration}{The 52x configuration}}}
\camera{For each four-TR subwindow and voxel slot, the bottom encoder applies a shared 1D temporal CNN: $1\to64$ channels with kernel 5/padding 2, GELU, then $64\to64$ with kernel 3/padding 1 and GELU. It averages over the subwindow's four time positions, concatenates the 16 slot embeddings, and projects $16\cdot64=1,024$ values to width 192. The output has shape $[B,12{,}358,8,192]$. Each of the three factorized encoders adds its own learned temporal-position table of shape $[1,1,8,192]$.}

\camera{Each transformer block is pre-layer-normalized attention with a residual connection, followed by a pre-layer-normalized feed-forward network $192\to768\to192$ with GELU and a residual connection. Attention has eight heads (head width 24) and dropout is zero. At a depth unit, spatial attention acts over all locations for each temporal token, followed by temporal attention over eight tokens at each location. Depths 3/2/1 therefore mean 6/4/2 attention blocks at chunk/parcel/network levels. These are full spatial attention operations, including 12,358 chunk positions; token compression does not imply a short encoder sequence or linear spatial-attention complexity.}

\camera{The three quantized arrays, excluding batch, have shapes $[12{,}358,8,18]$, $[1{,}024,8,12]$, and $[18,8,6]$. Their last dimensions are sign bits. BSQ entropy groups have sizes 6/4/3, giving 3/3/2 groups at the respective levels. No lookup table of $2^{18}$ learned vectors is needed to define the bottom quantizer. Learned and soft cross-attention pools add query arrays of size $1,024\times192$ and $18\times192$, shared across all temporal tokens and subjects. Normalized queries and normalized key/value features enter eight-head attention; there is no query-residual bypass around the pool output.}

\camera{For head $h$, the chunk-to-parcel pool uses weights:

\begin{equation}
\alpha^{(h)}_{pc}(w,\tau)=\operatorname{softmax}_{c}
\left(\frac{(W_Q^{(h)}\operatorname{LN}(a_p))^\top
W_K^{(h)}\operatorname{LN}(z_C[w,c,\tau])}{\sqrt{24}}\right)
\end{equation}

where $a_p$ is a learned query. The output is a projected concatenation of weighted value sums. Weights can change with subject, window, and time token through the input data, but no per-subject parameters are learned. Attention is dense and does not impose a one-parent assignment. There are no chunk/parcel/network spatial-position embeddings in this configuration. Within-chunk position enters through concatenation order, and temporal position enters through the position tables and CNN boundary effects.}

\camera{Let $\pi(p)$ be parcel $p$'s network, $\rho(c)$ chunk $c$'s parcel, and $(c(v),k(v))$ voxel $v$'s chunk and slot. For the temporal token covering time $t$, decoding is

\begin{align}
\hat y_N[n,t]&=H_N(r_N[n])_t,\\
\hat y_P[p,t]&=\hat y_N[\pi(p),t]+H_P(r_P[p])_t,\\
\hat y_C[c,t,k]&=\hat y_P[\rho(c),t]+H_C(r_C[c],\bar y_P[c],s_k)_t,\\
\hat x[v,t]&=\hat y_C[c(v),t,k(v)]
\label{eq:decoder}
\end{align}

Here $r_\ell=W_\ell^{\mathrm{post}}q_\ell$, $q_\ell=\operatorname{sign}(\bar u_\ell)/\sqrt{b_\ell}$, and $\bar u_\ell$ is the normalized linear projection of the encoder features. Temporal-token indices are suppressed.}

\camera{After quantization, each level projects back to width 192. The network and parcel heads are linear maps $192\to4$. For chunk $c$ and time token $\tau$, define:

\begin{equation}
\bar y_P[c,\tau]=\frac14\sum_{t=4\tau}^{4\tau+3}\hat y_P[\rho(c),t],
\quad r'_C[c,\tau]=W_C^{\mathrm{post}}q_C[c,\tau]+W_{\mathrm{cond}}\bar y_P[c,\tau]
\end{equation}

For each voxel slot $k$, a shared MLP $192\to192\to192\to4$ receives $r'_C[c,\tau]+s_k$ and produces the residual in Eq.~\eqref{eq:decoder}; $s_k$ is one of 16 learned 192-dimensional embeddings. Decoder heads have no attention layers. Parent broadcasts are the only cross-location coupling in decoding. No requirement forces voxel residuals to average to zero within parcels, or parcel residuals within networks. Direct coarse-head losses and coarse metrics on pooled voxel outputs are therefore distinct quantities.}

\subsection{\camera{\texorpdfstring{The 415$\times$ configuration}{The 415x configuration}}}
\camera{\hbarlow{} uses fixed pooling with one temporal token per 32-TR window at every level. Simply averaging the CNN features over all 32 TRs discarded much of their positional information in development. The final encoder instead learns eight weight vectors over the window positions, initialized to means of eight consecutive four-TR segments. The resulting $16\cdot64\cdot8=8,192$ features are projected to width 192 before the single bottom code. These eight continuous pooling slots are internal encoder features, not eight stored temporal codes.}

\camera{The middle head is a three-layer MLP $192\to192\to192\to32$ with a learned 192-dimensional parcel-slot embedding added after quantization. The top head and bottom slot decoder also output 32 TRs per code. The code dimensions are 18/12/24, and the middle entropy group uses all 12 bits jointly. Its configuration value \texttt{bsq\_mid\_inv\_temperature} is 0.3, versus 1 for the 52$\times$ configuration; this changes entropy regularization, not the hard sign code's bit count. The remaining encoder widths, attention heads, and level depths follow the 52$\times$ configuration. The final fixed models have 5,646,832 parameters at 52$\times$ and 7,313,174 at 415$\times$; the compact code does not imply fewer model parameters.}

\camera{The compact setting adds a direct middle-residual loss for stability:

\begin{equation}
10\,\MSE\!\left(H_P(r_P)[p,t],\ y_P[p,t]-y_N[\pi(p),t]\right).
\label{eq:midresid}
\end{equation}

The target uses the \emph{true} network signal, not the predicted network output. The predicted parcel signal is detached when used as the bottom decoder's base and conditioning input; voxel-loss gradients through these parent-prediction paths are stopped. This does not freeze the shared bottom encoder, which still feeds the higher levels. These changes in time representation, head capacity, and code allocation mean the 52$\times$--415$\times$ comparison measures two complete trained systems rather than the isolated effect of temporal token count.}

\section{\camera{Losses, optimization, and runtime}}
\label{app:training}

\subsection{\camera{Loss reductions}}

\camera{For each tensor, MSE and MAE denote means over its batch, time, and spatial elements. $\Delta$ is a first difference along the TR axis and has $W-1$ positions. The terms in Eq.~\eqref{eq:loss} are:

\begin{align}
\mathcal L_V&=\MSE(\hat x,x),&\mathcal L_{\Delta V}&=\MSE(\Delta\hat x,\Delta x),\\
\mathcal L_C&=\MAE(\hat y_C,y_C),&\mathcal L_{\Delta C}&=\MAE(\Delta\hat y_C,\Delta y_C),\\
\mathcal L_P&=\MSE(\hat y_P,y_P),&\mathcal L_N&=\MSE(\hat y_N,y_N).
\end{align}

Voxel losses exclude padded slots because the voxel signal is assembled from valid indices. Both chunk L1 terms include padded slots, whose targets are zero (3.7\% of chunk slots). They are not masked versions of the voxel losses. The parcel/network terms supervise the decoder's direct heads. Their targets are computed under the training grouping; the shuffled control consequently changes both aggregation and corresponding coarse supervision.}

\camera{BSQ regularization sums a commitment penalty and entropy terms at each level:

\begin{equation}
\mathcal L_{\mathrm{BSQ},\ell}
=\beta(s)\,\MSE(\bar u_\ell,\sg[q_\ell])
+\frac{\zeta(s)}{\eta_\ell}\bigl(H_{\mathrm{sample},\ell}-H_{\mathrm{code},\ell}\bigr).
\end{equation}

Here $s$ counts microbatches per rank and $\eta_\ell$ is the implementation's inverse-temperature setting. Both entropy coefficients inside the parentheses are 1. The per-sample entropy is computed analytically from the sign probabilities; code-distribution entropy uses the configured bit groups and aggregates token probabilities within the batch. It is a regularizer on soft assignments, not an entropy-coded payload estimate. Hard-code perplexity is a diagnostic of the empirical code distribution, not the number of bits needed to serialize the fixed-length representation.}

\begin{table}[ht]
\centering

\caption{\camera{Final hierarchical training recipes. All listed weights are constant except the BSQ ramps described below. A dash means the term is inactive.}}
\label{tab:training}

\camera{\small
\begin{tabular}{lrrr}
\toprule
Setting & Fixed / learned / shuffled 52 & Soft 52 & Fixed 415 \\
\midrule
Epochs & 10 & 10 & 10 \\
Learning rate & $2\cdot10^{-4}$ & $2\cdot10^{-4}$ & $2\cdot10^{-4}$ \\
Effective batch & 32 & 32 & 32 \\
$\lambda_V,\lambda_{\Delta V}$ & $1,0.5$ & $1,0.5$ & $1,0.5$ \\
$\lambda_C,\lambda_{\Delta C}$ & $10,2$ & $10,2$ & $10,2$ \\
$\lambda_P,\lambda_N$ & $10,10$ & $10,10$ & $10,10$ \\
BSQ coefficient & 0.03 & 0.03 & 0.03 \\
Alignment coefficient ($\sigma$) & -- & 2 & -- \\
Direct middle-residual coefficient & -- & -- & 10 \\
\bottomrule
\end{tabular}}
\end{table}

\subsection{\camera{Optimization and checkpoint selection}}
\camera{Hierarchical runs use AdamW, weight decay 0.01, gradient-norm clipping at 1, and bf16 automatic mixed precision on NVIDIA H100 80\,GB GPUs. The learning rate is constant; there is no learning-rate warmup or scheduler. The final hierarchy runs use one GPU, batch size 4, and gradient accumulation of 8 (effective batch 32), with shuffled training batches. Commitment $\beta$ ramps linearly from 0.01 to 0.1 and entropy weight $\zeta$ from 0.1 to 1 over the first 4,000 microbatches. On one GPU this is 500 optimizer updates. The final epoch, index 9, is used for every primary reconstruction and tokenizer readout. Seed labels 0, 1, and 2 identify independent training runs, not bootstrap draws.}

\subsection{\camera{Flat baseline settings}}

\camera{The flat BSQ baselines remove the parcel and network levels from HBAR and keep only its chunk-level encoder and decoder (same encoder architecture, 3-layer reconstruction head, and BSQ quantizer with L2 normalization and analytical entropy regularization). Each chunk token is quantized to 19 sign bits; because 19 is prime, the codebook-entropy term is computed bit-wise rather than over bit groups. Rates are matched to HBAR by the number of time tokens per chunk: flat BSQ-52 uses 8 tokens per chunk per 32-TR window (12{,}358 $\times$ 8 $\times$ 19 = 1{,}878{,}416 bits/window, vs.\ 1{,}878{,}720 for HBAR-52), and flat BSQ-415 pools the window to one token per chunk using the same temporal pooling as HBAR-415 (12{,}358 $\times$ 19 = 234{,}802 bits/window, vs.\ 235{,}164). The models have 2{,}965{,}719 (52$\times$) and 4{,}346{,}291 (415$\times$) parameters, vs.\ 5{,}646{,}832 and 7{,}313{,}174 for HBAR. The objective is HBAR's without the coarse-scale terms: voxel reconstruction MSE (weight 1), voxel temporal-difference MSE (0.5), chunk reconstruction L1 (10), chunk temporal-difference L1 (2), and the BSQ term (0.03); the parcel and network reconstruction losses do not exist at this depth. Training matches HBAR: 10 epochs, learning rate $2\times10^{-4}$, bf16 mixed precision, effective batch 32 (one GPU, batch 4, gradient accumulation 8), on the same 877 training subjects. The flat baselines are therefore rate- and recipe-matched to HBAR, but not parameter-matched, and differ from it by the absence of the parcel and network levels and their losses.}

\subsection{\camera{Cost and computational scope}}
\camera{Ten-epoch 52$\times$ hierarchical runs require approximately 36--38 H100 GPU-hours; representative peak PyTorch allocations are about 56--57\,GB per GPU at batch 4, with roughly 70\,GB reserved. Flat baselines use about 35 GPU-hours at $52\times$. These are indicative training costs, and several factors can affect elapsed time. 415$\times$ runs completed in approximately 14 single-GPU hours. No matched final-model wall-clock-time or inference-speed comparison is claimed.}

\subsection{Loss-component ablations}
\label{app:loss-ablation}

To isolate the role of the three auxiliary terms in Eq.~\eqref{eq:loss}, we ran single-knob ablations during development: each changes one coefficient of the 52$\times$ recipe and holds everything else fixed, and each is compared with its own reference run at a matched epoch. These are single-seed development runs on validation subjects, not the final three-seed test-set models of Table~\ref{tab:recon}, so the absolute values are not comparable across rows; the within-row comparison is the quantity of interest. Reconstruction is a voxel-level MSE, and atlas latent MSE is the squared distance between the model's pooled parcel/network features and the atlas-pooled targets, evaluated on the same windows. Code usage and perplexity are measured on the parcel-level (middle) quantizer.

\begin{table}[h]
\centering
\small
\setlength{\tabcolsep}{4pt}
\begin{tabular}{lcccc}
\toprule
Ablation & Recon. MSE$\downarrow$ & Atlas MSE$\downarrow$ & Usage$\uparrow$ & PPL$\uparrow$ \\
\midrule
$\lambda_{\mathrm{vq}}: 0.03\to0.003$
  & 0.668 / 0.670 & -- & 0.100 / 0.208 & 24.7 / 243.5 \\
$\lambda_{\Delta V},\lambda_{\Delta C}: (0.5,2)\to(0,0)$
  & 0.673 / 0.810 & 20.8 / 57.8 & -- & -- \\
$\sigma: 2\to0.5$
  & 0.779 / 0.733 & 40.7 / 66.0 & -- & -- \\
\bottomrule
\end{tabular}
\vspace{3mm}
\caption{Single-knob ablations at $52\times$. Values are reference/ablated at matched epochs. The first two rows use fixed-hierarchy references; the third uses a soft-hierarchy reference. Rows are not directly comparable. Usage is the fraction of $2^{12}$ parcel codes observed; PPL denotes hard-code perplexity. Dashes indicate unmeasured quantities.}
\label{tab:loss-ablation}
\end{table}

Reducing the quantization weight tenfold leaves reconstruction essentially unchanged but roughly doubles parcel-level code usage and raises perplexity by an order of magnitude, so this term mainly governs code utilization rather than fidelity. Removing the temporal-difference terms is clearly harmful: reconstruction MSE rises by 20\% and the alignment distance to atlas-pooled features nearly triples, so $\mathcal{L}_\Delta$ shapes the latent structure as well as temporal fidelity. Weakening the alignment weight from $\sigma = 2$ to $0.5$ exposes the intended trade-off: reconstruction MSE falls by 6\% while atlas latent MSE rises by 62\%, i.e., the soft hierarchy buys reconstruction by moving away from the atlas, which is the tunable behavior it was designed to provide.

\section{\camera{Complete bit accounting}}
\label{app:bits}

\camera{The numerator is the preprocessed masked signal: 32 TRs, 190,468 unique voxels, 16 bits/value. It is not the full rectangular image, raw scanner data, or a float32 array. Padding is not counted as input data. Every stored latent code, including empty parcel slots and the unassigned network group, is counted in the denominator.}

\camera{\begin{align}
B_{52}&=12{,}358\cdot8\cdot18+1{,}024\cdot8\cdot12+18\cdot8\cdot6\\
&=1{,}779{,}552+98{,}304+864=1{,}878{,}720,\nonumber\\
B_{415}&=12{,}358\cdot18+1{,}024\cdot12+18\cdot24\\
&=222{,}444+12{,}288+432=235{,}164.\nonumber
\end{align}}

\camera{Flat BSQ needs 19 bits per chunk, giving 234,802 bits/window. Flat models at $415\times$/$52\times$ emit 1/8 codes per chunk per 32-TR window.}

\begin{table}[ht]
\centering
\caption{\camera{Compression accounting. ``Window ratio'' uses one 32-TR window. Scan ratios use a 1,200-TR scan: 38 nonoverlapping padded windows (stride 32) or 74 overlapping windows (stride 16).}}
\label{tab:bits}
\setlength{\tabcolsep}{4pt}
\camera{\small
\begin{tabular}{lrrrrr}
\toprule    
Model & Bits/window & Window ratio & Bits/sample & Tiled scan & Stride-16 scan \\
\midrule
HBAR-52      & 1,878,720 & 51.91  & 0.30824 & 51.22  & 26.30 \\
Flat BSQ-52  & 1,878,416 & 51.92  & 0.30819 & 51.23  & 26.31 \\
\midrule
HBAR-415     & 235,164   & 414.69 & 0.03858 & 409.23 & 210.15 \\
Flat BSQ-415 & 234,802   & 415.33 & 0.03852 & 409.86 & 210.47 \\
Flat RVQ-415 & 234,802   & 415.33 & 0.03852 & 409.86 & 210.47 \\
MS-VQVAE-415 & 234,732   & 415.45 & 0.03851 & 409.98 & 210.53 \\
\bottomrule
\end{tabular}}
\end{table}

\camera{The scan calculation is $R_{\mathrm{scan}}=1,200\cdot V\cdot16/(n_w B_q)$. The rates assume ideal fixed-length bit packing with no entropy coding. For $M$ windows, shared storage $S$ bits, and per-window/container overhead $H$ bits, a model-inclusive ratio is

\begin{equation}
R_{\mathrm{total}}(M)=\frac{M B_x}{M(B_q+H)+S}
\end{equation}

The recorded layout arrays occupy 1,041,946 bytes before container overhead; a full float32 weight array occupies $4P$ bytes for $P$ parameters (about 22.6\,MB for fixed 52$\times$ and 29.3\,MB for fixed 415$\times$). Which encoder/decoder weights are distributed changes $S$. Finally, the 36- or 54-dimensional probe vector is a mean-pooled readout of many discrete tokens. It is not the complete encoded window and is not sufficient for the reconstruction decoder. Comparing 36 readout coordinates with 1,024 DiFuMo coordinates is a dimensionality comparison, not a 28-fold codec advantage. Quantization bounds the code alphabet; it does not guarantee predictive sufficiency or privacy.}

\section{\camera{Reconstruction metrics and statistical analysis}}
\label{app:metrics}

\camera{For each level, a window's correlation is Pearson's $r$ between vectors formed by flattening time and units. It is not the mean of independent per-voxel temporal correlations. Window scores are averaged within each subject, then equally across subjects. MSE is likewise computed per window then averaged by subject. There is no Fisher $z$ transform. With $J=16$ windows for subject $i$, the reported mean of metric $m$ is

\begin{equation}
\bar m=\frac1{109}\sum_{i=1}^{109}\left(\frac1{16}\sum_{j=1}^{16}m_{ij}\right)
\end{equation}

Windows from the same subject are not treated as independent observations for confidence intervals.}

\camera{All coarse reconstructions are formed from valid voxel slots using Eq.~\eqref{eq:truepool}, not the model's direct coarse decoder heads. In particular, each parcel prediction sums valid decoded voxels and divides by that parcel's valid voxel count. A reconstruction of the true chunk inputs must reproduce the dataset's parcel/network targets within float16 tolerance. A 300-window oracle check gave parcel MSE $3.1\cdot10^{-9}$ and network MSE $8.2\cdot10^{-10}$, with least-squares scales 1.000000 and 1.000002. The evaluator checks MSE below $10^{-7}$, maximum absolute error below $4\cdot10^{-3}$, scale error below $10^{-4}$, and zero targets for empty parcels. A correlation-only check would not detect a multiplicative scale error.}

\camera{Parcel MSE averages over all 1,024 slots, including 14 zero-error empty parcels. Multiplying by $1,024/1,010=1.01386$ converts it to the mean over nonempty parcels. Correlations include the same zero slots for every model; they cannot be converted by that MSE factor. The final 18-group network metric includes the unassigned group. Coarse-scale correlation is consequently atlas- and convention-dependent.}

\camera{Reconstruction CIs use 10,000 subject-bootstrap resamples; paired contrasts resample the same subjects for both models. The accompanying two-sided paired sign-flip test uses 10,000 draws and a plus-one correction, with minimum attainable $p=1/10,001$. Many contrasts, including those between fixed-model seeds, reach this floor. They quantify consistency across these subjects for trained checkpoints. We emphasize effect sizes, paired CIs, and separately reported seed spread.

Paired tests show HBAR better at parcel scale than every comparator at every seed. At $415\times$, MS-VQVAE achieves the best voxel reconstruction across all seeds, while HBAR dominates for parcel and network reconstruction. At $52\times$, atlas-consistent grouping improved parcel correlation relative to shuffled grouping in all three seed pairs ($\Delta r=0.029$--$0.040$). Network correlation, by contrast, was similar on average, and the direction of the difference varied across seeds ($+0.013$, $-0.006$, and $-0.008$).}

\section{\camera{Downstream features, baselines, and probes}}
\label{app:probes}

\paragraph{Linear probes} \camera{use the same stride-64 windows for every method: 20,808 windows total, of which 16,651 belong to training subjects. This differs intentionally from the denser stride-16 training index and from the 16-window-per-subject reconstruction sample. A window's hierarchical feature concatenates the spatial/temporal means of its three post-quantization sign vectors; a subject's feature averages these window vectors. Dimensions are $18+12+6=36$ for 52$\times$ and $18+12+24=54$ for 415$\times$. Flat BSQ has 19 coordinates. The readout discards position and order and need not exploit all information present in the stored sequence.}

\paragraph{Autoregressive readout.} The AR model uses only finest-level codes, excluding MS-VQVAE's top-level tokens. HBAR's 18-bit codes are split into three 6-bit IDs (vocabulary 64); flat BSQ's 19-bit codes into 10- and 9-bit IDs (vocabulary 1,536); and RVQ/MS-VQVAE use their two codebook indices with offsets (vocabularies 1,536/1,024). IDs are ordered by temporal slot, chunk, and bit group. Non-overlapping 32-TR windows are concatenated chronologically within subjects. Each Transformer position embeds 64 consecutive IDs as a 256-dimensional vector (64$\times$4), with learned absolute positional embeddings. A 4-layer, 4-head causal pre-norm Transformer (width 256, FFN 1,024, GELU, dropout 0.1) predicts all 64 IDs of the next patch using separate softmax heads and mean cross-entropy. Samples span four windows (128 TRs), truncated to complete patches. Sequence lengths and parameter counts (4.9M--31.7M) are given in Table~\ref{tab:ar_seq}.

AR models are trained on 877 training subjects for 10 epochs (seed 0) using AdamW (LR $5\times10^{-5}$, weight decay 0.01, batch size 8, gradient clipping 1.0, bf16), with 200 warmup steps followed by cosine decay to 20\% of peak LR after the first epoch. Four-window samples are shuffled each epoch; the final checkpoint is used without validation-based selection. The frozen model processes each subject's sequence in segments of up to four windows; final-layer hidden states are averaged within and across segments to yield a 256-dimensional subject representation. These features are evaluated using the same standardized, balanced L2 logistic probe as above, with five-fold cross-validation for $C$ and 10,000 subject-bootstrap resamples. Probes are fitted on 877 training subjects and evaluated on 109 held-out test subjects. Although training-subject features come from sequences seen during AR training, phenotype labels are never used by the AR model. Table~\ref{tab:downstream} reports tokenizer and AR seed 0.

\begin{table}[h]
  \centering
  \small
  \caption{\camera{AR input sequences per tokenizer. IDs/window counts finest-level IDs per 32-TR window; positions are 64-ID patches per four-window sample. Parameter counts include embeddings and the output
  head.}}
  \label{tab:ar_seq}
  \begin{tabular}{lrrrr}
  \toprule
  Tokenizer & Vocab. & IDs / window & Positions & AR params \\
  \midrule
  HBAR-52 (fixed, learned, soft, shuffled) & 64    & 296,592 & 18,537 & 9.02M \\
  Flat BSQ-52                & 1,536 & 197,728 & 12,358 & 31.66M \\
  HBAR-415 (fixed)                         & 64    & 37,074  & 2,317  & 4.87M \\
  Flat BSQ-415 / Flat RVQ-415              & 1,536 & 24,716  & 1,544  & 28.89M \\
  MS-VQVAE-415                             & 1,024 & 24,716  & 1,544  & 20.47M \\
  \bottomrule
  \end{tabular}
  \end{table}

\camera{The atlas baselines use the same preprocessed signal and the same 20,808 stride-64 windows. DiFuMo time means average each of the 1,024 parcel series over the window; Yeo time means do the same for 18 groups (the 17 Yeo networks plus one group of parcels without a Yeo assignment). The parcel-series PCA baseline flattens each $32\times1{,}024$ window and projects it to 256 coordinates. Parcel-FC computes Pearson correlations between parcel series within each window, vectorizes the upper triangle ($1{,}024\cdot1{,}023/2=523{,}776$ entries), and projects it to 256 coordinates. Both PCA bases are fit only to the 16,651 training-subject windows. Features are then averaged by subject. Neither validation nor test windows are used to fit PCA or feature standardization.}

\camera{For classification, a training-set standardizer precedes an L2-regularized logistic regression with balanced class weights. Regularization $C$ is selected by five-fold inner cross-validation on training subjects from $\{10^{-5},10^{-4},10^{-3},10^{-2},10^{-1},1,10,100\}$. We report AUROC for sex and macro-averaged one-vs-rest AUROC for the three age bins. There are 877 training and 109 test sex labels; class counts are 480/397 in training and 58/51 in test. Age has 865 training labels and all 109 test labels, with test bin counts 19/59/31. Numeric representatives 23, 28, and 33 encode the age categories; this is age-bin classification, not fine-grained continuous-age prediction.}

\camera{Single-model CIs are 95\% percentile intervals from 10,000 subject-bootstrap resamples of the test set, conditional on the trained tokenizer and probe; they reflect subject sampling, not training-seed variance. Paired differences use 10,000 bootstrap resamples with the same subject indices for both prediction sets, and a two-sided permutation test that randomly swaps the two models' predicted probabilities within each subject (10,000 permutations, add-one smoothed). We do not interpret nonsignificance as equivalence.}

\section{\camera{Frequency-resolved reconstruction}}
\label{app:spectral}

\camera{For each 32-TR window, the evaluator uses an undetrended raw periodogram to compare Fourier power of the reconstruction residual with input power in broad frequency bands. For band $\mathcal B$, the error ratio is of the form

\begin{equation}
E_{\mathcal B}=\frac{\sum_{v}\sum_{f\in\mathcal B}|\mathcal F(\hat x-x)[v,f]|^2}
{\sum_{v}\sum_{f\in\mathcal B}|\mathcal F(x)[v,f]|^2}
\end{equation}

The exported window statistics are averaged within subjects and then across subjects, as for reconstruction. Zero indicates perfect reconstruction in the band; one indicates residual power equal to input power and need not imply statistical independence. Values can exceed one. It is not an explained-variance statistic and the bands are not validated partitions into neuronal, respiratory, and noise components.}

\begin{table}[ht]
\centering
\caption{\camera{Final test voxel spectral-error/input-power ratios (lower is better). Mean $\pm$ SD over three training seeds, rounded to two decimals. The DC column contains only the zero-frequency bin at this window length. SDs shown as 0.00 are below 0.005.}}
\label{tab:spectral}
\setlength{\tabcolsep}{3pt}
\camera{\small
\begin{tabular}{lrrrrr}
\toprule
Model & DC & 0.02--0.10 & 0.10--0.20 & 0.20--0.35 & 0.35--0.70 Hz \\
\midrule
HBAR-52 (fixed) & $0.38\pm0.02$ & $0.66\pm0.02$ & $0.85\pm0.01$ & $0.94\pm0.01$ & $0.90\pm0.00$ \\
Learned pooling (52$\times$) & $0.42\pm0.02$ & $0.68\pm0.02$ & $0.85\pm0.01$ & $0.92\pm0.00$ & $0.88\pm0.01$ \\
Soft pooling (52$\times$) & $0.45\pm0.23$ & $0.64\pm0.08$ & $0.81\pm0.04$ & $0.94\pm0.01$ & $0.99\pm0.01$ \\
Shuffled grouping (52$\times$) & $0.40\pm0.02$ & $0.66\pm0.02$ & $0.84\pm0.01$ & $0.92\pm0.01$ & $0.89\pm0.01$ \\
Flat BSQ-52 & $0.37\pm0.03$ & $0.68\pm0.01$ & $0.86\pm0.00$ & $0.93\pm0.01$ & $0.88\pm0.02$ \\
\midrule
HBAR-415 (fixed) & $0.41\pm0.01$ & $0.94\pm0.00$ & $0.99\pm0.00$ & $1.00\pm0.00$ & $1.00\pm0.00$ \\
Flat BSQ-415 & $0.42\pm0.01$ & $0.97\pm0.00$ & $1.00\pm0.00$ & $1.00\pm0.00$ & $1.00\pm0.00$ \\
Flat RVQ-415 & $0.39\pm0.03$ & $0.94\pm0.00$ & $1.00\pm0.00$ & $1.00\pm0.00$ & $0.99\pm0.00$ \\
MS-VQVAE-415 & $0.29\pm0.00$ & $0.99\pm0.00$ & $1.00\pm0.00$ & $1.00\pm0.00$ & $1.00\pm0.00$ \\
\midrule
\bottomrule
\end{tabular}}
\end{table}

\camera{With $W=32$ and TR $0.72$\,s, the frequency spacing is $1/(32\cdot0.72)=0.0434$\,Hz and Nyquist is $0.6944$\,Hz. The $0$--$0.02$\,Hz band contains only DC; the $0.02$--$0.10$ and $0.10$--$0.20$\,Hz bands
contain two positive-frequency bins each, $0.20$--$0.35$\,Hz four and $0.35$--$0.70$\,Hz eight. At 52$\times$, all models recover part of the within-window dynamics below 0.20\,Hz, and HBAR-52 is comparable to the other 52$\times$ variants. At 415$\times$, every model has one temporal token per chunk and window, and its residual carries essentially all input power above 0.10\,Hz (ratios $\geq 0.987$). Differences are confined to the two 0.02--0.10\,Hz bins, where HBAR-415 and flat RVQ-415 (0.94) retain slightly more than flat BSQ-415 (0.97) and MS-VQVAE (0.99), and to the DC bin, where MS-VQVAE is lowest. We therefore avoid claims about resolving narrow slow bands or preserving complete resting-state dynamics. Strong flattened coarse correlations can coexist with substantial within-window temporal error. A long-scan spectral or connectivity-preservation study would answer a different question and is not established by this table.}

% END APPENDICES

%%%%%%%%%%%%%%%%%%%%%%%%%%%%%%%%%%%%%%%%%%%%%%%%%%%%%%%%%%%%

% \newpage
% \input{checklist.tex}

\end{document}